%% file: main.tex
\documentclass{article}

\usepackage{stylo}

\title{Trapping LLM \enquote{Hallucinations} Using Tagged Context Prompts}
\author[1, 2]{Philip Feldman}
\author[2]{James R. Foulds}
\author[2]{Shimei Pan}

\affil[1]{ASRC Federal}
\affil[2]{University of Maryland, Baltimore County}

\begin{document}

\maketitle
\begin{abstract}
    Recent advances in large language models (LLMs), such as ChatGPT, have led to highly sophisticated conversation agents. However, these models suffer from \textit{hallucinations,} where the model generates false or fabricated information. Addressing this challenge is crucial, particularly with AI-driven platforms being adopted across various sectors. In this paper, we propose a novel method to recognize and flag instances when LLMs perform outside their domain knowledge, and ensuring users receive accurate information.
    
    We find that the use of context combined with embedded tags can successfully combat hallucinations within generative language models. To do this, we baseline hallucination frequency in no-context prompt-response pairs using generated URLs as easily-tested indicators of fabricated data. We observed a significant reduction in overall hallucination when context was supplied along with question prompts for tested generative engines. Lastly, we evaluated how placing tags within contexts impacted model responses and were able to eliminate hallucinations in responses with 98.88\% effectiveness.
\end{abstract}

\input{text/introduction}

\input{text/background}
\input{text/methods}
\input{text/results}

\input{text/conclusions}

\newpage
\bibliographystyle{plain}

\end{document}

%% file: text/introduction.tex
\section{Introduction}

\begin{figure}[htb]
    \centering
    \fbox{\includegraphics[width=0.65\textwidth]{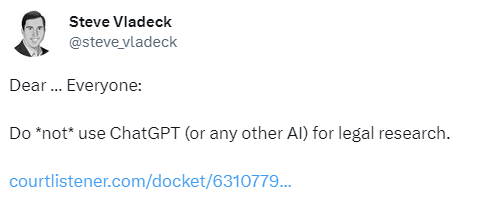}}
    \caption{Naive acceptance of LLM \enquote{hallucination} is becoming widespread}
    \label{fig:tweet}
\end{figure}

Recent advances in large language models (LLMs), such as ChatGPT, have revolutionized the field of artificial intelligence, leading to highly sophisticated conversation agents capable of generating coherent and context-appropriate responses. However, despite their remarkable performance, these LLMs suffer from a critical issue: hallucinations. This phenomenon, where the model generates false or fabricated information, poses a significant challenge, especially when users trust and rely on these systems for critical decision-making or research. Addressing this issue is of paramount importance, especially considering the widespread adoption of AI-driven platforms across various sectors, including legal, medical, and educational fields.

An alarming example of the implications resulted from a recent court case where the defense attorney used ChatGPT to performlegal research, and unknowingly cited non-existent cases\footnote{See \surl{https://storage.courtlistener.com/recap/gov.uscourts.nysd.575368/gov.uscourts.nysd.575368.32.1_1.pdf}}. Though the court did not find malpractice on this occasion, it highlighted the urgent need to curtail hallucinations in LLMs to prevent such issues in the future. Naturally, the news of such gullible behavior wuickly made the rounds of social media and was picked up by major newspapers such as the New York Times\footnote{\surl{https://www.nytimes.com/2023/05/27/nyregion/avianca-airline-lawsuit-chatgpt.html}}.

Training users to adopt a critical thinking approach towards the output of LLMs may seem like the most straightforward solution, but it is far removed from the reality of how humans interact with, and trust, these AI systems. Consequently, it is essential to develop AI models capable of generating reliable and accurate responses, with built-in mechanisms to detect potential hallucinations and warn users if responses fall outside the realm of known information.

In this paper, we aim to tackle the prevalent issue of hallucinations in LLMs, focusing on constraining models to produce reliable and trustworthy output. We propose a novel method to recognize instances where the model performs outside its domain knowledge and suggest mechanisms to flag such outputs as potential hallucinations. Our approach offers a significant improvement in mitigating the risks associated with these powerful AI systems while ensuring that users receive accurate and reliable information to make informed decisions across various professional fields.

%% file: text/background.tex
\section{Background}
The recent advancements in Large Language Models (LLMs) have unlocked a new range of opportunities, with models like GPT-3 exhibiting skilled performance in various NLP tasks~\cite{bang2023multitask}. However, alongside these benefits, LLMs, including ChatGPT, are known to hallucinate facts and generate non-factual statements that could impact the trustworthiness of their outputs~\cite{manakul2023selfcheckgpt}.  The challenge of fact-checking and mitigating hallucinations in LLMs has become increasingly vital for maintaining reliability and user trust.

Validating the trustworthiness of the output of neural networks is an ongoing problem. Deep Mind's AlphaGo used a combination of Monte Carlo tree search, deep neural networks, and reinforcement learning, leading to the creation of separate policy and value networks that enabled determining the best move in the game of Go~\cite{holcomb2018overview}. Drawing parallels with this approach, it is conceivable that the evaluation of generated text can incorporate a separate validation model that carefully examines the generated outputs.

However, just as AlphaGo's success is not without its limitations, applying a similar approach to generated text analysis could face challenges as well. Hallucinations may evade detection by the validation model, as they might conform to the syntactic and semantic rules expected within the generated text. Moreover, the separate evaluation model might suffer from signal bias, wherein it tends to endorse certain patterns, content, or styles over others, exacerbating the risk of hallucinations and resulting in partially validated outputs that still lack contextual or factual integrity.

The investigation of ethical risks and harmful consequences in using LLMs, such as biases, reliability, robustness, and toxicity, emphasizes the need for practical and responsible LLM designs~\cite{zhuo2023exploring}. Techniques such as SelfCheckGPT have been proposed to make use of the stochasticity in sampled responses to assess the factuality of generated outputs~\cite{manakul2023selfcheckgpt}. Such approaches have shown promise in detecting non-factual and factual sentences and ranking passages based on factuality.

Furthermore, studies have analyzed the shortcomings and limitations of LLMs like ChatGPT, highlighting the possible failures across a range of issues including reasoning, factual errors, and bias~\cite{borji2023categorical}. Modeling these limitations and understanding their implications remain significant challenges for future improvements in LLMs and their usage in practical applications. 

Behavioral studies, such as those performed by McKenna et al.~\cite{mckenna2023sources}, have established factors that predict LLMs performance. Their findings suggest that memorization of training data and utilizing corpus-based heuristics using relative word frequencies contribute to hallucination issues in generative models. Despite these insights, addressing the hallucinations and classification vulnerabilities remains a critical area of exploration.

Additionally, LLMs may be exploited through the use of \enquote{jailbreak} prompts. Jailbreaking is a form of prompt engineering that seeks to modify or manipulate an AI model's behavior or output beyond its intended capabilities. It involves exploiting vulnerabilities in the model's programming or configuration to remove limits or restrictions that were built in by the developers. This may allow the model to generate new and unexpected outputs, or even to act in ways that were not intended or predicted by its creators. Jailbreak prompts can consistently evade the restrictions devised by their creators~\cite{liu2023jailbreaking}.

There remains a gap in the literature regarding how the use of context prompts can reduce hallucination problems in LLMs, ensuring both the appropriate outputs and preserving the comprehensibility of the generated responses. As the potential for toxic output has been noted in unpredictable contexts~\cite{zhuo2023exploring}, the ability to verifiably prevent hallucination outputs is essential to ensuring reliability and trustworthiness in LLMs. We will describe how the use of tagged context prompts can be used as an effective tool in reaching this goal.

%% file: text/methods.tex
\section{Methods}

In our study, we employed a multi-stage approach to assess the impact of tagged context prompts on the hallucination problems in LLMs, with a focus on the OpenAI GPT-3 and later models. The methods included generating question sets, creating context prompts through summaries, verifying context prompts and questions, and performing experiments with different GPT models using context-based and context-free questions. We implemented the following procedures:

\paragraph{Data Set Creation} A crucial aspect of our study was creating a diverse and comprehensive set of questions and contexts to examine the abilities of generative language models, particularly OpenAI's GPT-3 and its successors. We implemented a custom script to generate these questions based on multiple contexts. These contexts comprised summarized Wikipedia articles on various subjects, as well as sections from the lead author's recent book, published in May 2023. The questions aimed to cover a wide range of topics, from social issues and scientific matters to technical domains. This approach was designed to thoroughly assess the based model's ability to answer questions in various domains when using a context and when no context was provided.

To generate these questions, the script extracted contextual information from the following sources:
\begin{itemize}[noitemsep]
    \item Summarized Wikipedia articles on subjects such as France, chess, the solar system, dating, continents, and the novel \enquote{To Kill a Mockingbird.}
    \item Selected sections from the lead author's book on computational sociology, published in May 2023~\cite{feldman2023stampede}, which would not have included in any GPT training data.
\end{itemize}

As a result, our question set included diverse examples such as:

\begin{itemize}[noitemsep]
    \item (France) What is the capital of France?
    \item (Chess) In the game of chess, which piece can only move diagonally?
    \item (Solar System) How does the distance between each planet's orbit and the next nearest object to the Sun generally increase?
    \item (Dating) What precautions does Sara McCorquodale advise women to take when meeting strangers on dates?
    \item (Continents) What are the four attributes geologists use to define a continent?
    \item (To Kill a Mockingbird) How has the reception of 'To Kill a Mockingbird' changed over time?
    \item (Author's Book) How can technology be utilized to effectively introduce diversity injection techniques into various social networks?
\end{itemize}

By including topics like \textit{dating}, for which the LLM would have received extensive training data, and others like the author's computational sociology text, representing a smaller training domain, we aimed to provide a more controlled evaluation of the LLM performance in and out of its trained timeframes.

\paragraph{Verification} To maintain the reliability and accuracy of our research, we implemented a meticulous verification process. This process focussed on guaranteeing that the context prompts were a precise reflection of the source material, particularly the book sections, and that the generated questions were coherent, relevant, and strongly linked to the given context.

The verification process comprised the following steps:

\begin{enumerate}[noitemsep]
    \item \textbf{Cross-referencing}: We cross-referenced context prompts with their original sources, ensuring their accuracy and completeness. This step was vital to avoid any inaccuracies that might have arisen during context generation from the summarized Wikipedia articles and the lead author's book.
    
    \item \textbf{Relevance assessment}: We examined the generated questions to confirm that they were pertinent to the context prompts. This assessment also involved evaluating whether the questions exemplified the corresponding subjects well, such as France, chess, solar system, dating, continents, \enquote{To Kill a Mockingbird,} and the author's book on computational sociology.
    
    \item \textbf{Coherence check}: We thoroughly inspected the questions for logical coherence and linguistic correctness. This step ensured that the generated questions made sense and were grammatically sound, enabling a meaningful evaluation of the generative language models' capabilities.
\end{enumerate}

By conducting this comprehensive verification process, we minimized the risk of false conclusions deriving from inaccurate context prompts or poorly formulated questions. 

\paragraph{Tag Placement} For the purpose of examining the impact of tagged context prompts on generative language models' responses, we developed a program that embedded tags within the context prompts generated earlier. These tags were designed in the form of \enquote{(source x)}, where \enquote{x} represented a unique identifier for the source, and helped guide the language models in answering questions. We had discovered through previous experimentation that prompts to the language models that ended with \enquote{Provide details and include sources in the answer} would reference these tags if present or attempt to create citations to sources as part of the response.

Our program employed the following process for tag placement:

\begin{enumerate}[noitemsep]
    \item \textbf{Unique Tag Creation}: We generated unique 4-digit numbers to serve as identifiers for each source. For instance, a tag like \enquote{(source 3626)} would uniquely correspond with a specific line in the source material.
    
    \item \textbf{Tag Insertion}: The program inserted the generated tags, such as \enquote{(source 3626)}, within the context prompts at the end of each sentence. By incorporating these tags into the context, we would provide the needed contextual hint for the generative language models.
\end{enumerate}

An example snippet of a tagged context is displayed below:

\begin{verbatim}
    ...fill the same needs as belonging to a cult(source 3626). 
    It works one person at a time, nudging them...
\end{verbatim}

Through the use of tags, we were able to systematically analyze the role of these additional context elements in influencing the language models' responses, while providing further hints for them to generate accurate and contextually relevant answers.

\paragraph{Experiments} To thoroughly assess the impact of the tagged context prompts on question answering using generative language models, we conducted various experiments involving multiple permutations of context, question, and engine. This comprehensive approach allowed us to investigate the influence of inserting tags into source contexts on the generation of hallucinated information and the LLMs' ability to validate responses using random \enquote{known good} tags that could not be synthesized by the models otherwise.

Each experiment iteration included the following components:

\begin{enumerate}[noitemsep]
    \item \textbf{No-context Question}: We used one of the pre-generated questions which did not include any supporting context. To encourage the model to provide citations, we appended the instruction \enquote{Provide details and include sources in the answer.} to the question.
    
    \item \textbf{Tagged-context Question}: We incorporated the pre-generated context tagged as described earlier. To ensure a comprehensive evaluation, each context was applied across all questions, making it relevant for one group of questions and not relevant for the others.
    
    \item \textbf{Generative Language Model Selection}: We tested the prompts on a variety of recent generative language models, including GPT-4, GPT-4-0314, GPT-3.5-turbo, GPT-3.5-turbo-301, text-003-davinci, davinci-instruct-beta, and curie-instruct-beta. The results of each experiment were stored for further analysis.
\end{enumerate}

By conducting these extensive experiments, we could systematically analyze the effectiveness of our tagged-context approach, its implications on context-awareness in large language models, and the tag-driven validation of their responses. 

\paragraph{Data Collection} To systematically analyze the outcomes of our experiments, we stored various data components in structured JSON files. Each file consisted of detailed information, such as the context, question, context-free response, context-based response, and referenced source IDs for each model. This approach allowed us to effectively track and assess the performance of different LLMs with respect to our tagged-context technique and its impact on the reduction of hallucinations and improvement of generated responses.

In addition to storing the raw context-based responses, we applied a regular expression (regex) to \enquote{clean} the responses by removing the tags. This allowed us to examine the feasibility of presenting the generated responses to end users without the distracting tags. For instance:

Raw response:
\begin{verbatim}
...nudging them off the stampede trajectory (source 3626). Social media...
\end{verbatim}

Cleaned response:
\begin{verbatim}
...nudging them off the stampede trajectory. Social media...
\end{verbatim}

During the rest of this paper, we will be showing unprocessed text strings with sources visible for clarity.

To facilitate analysis, we processed the JSON files and transformed them into Excel spreadsheets. This helped in efficiently evaluating the results of our experiments and drawing insights into the effectiveness of our tagged-context approach.

All the relevant code supporting our experiments can be found on GitHub at \surl{https://github.com/pgfeldman/KeywordExplorer/tree/main/keyword_explorer/experiments}, while the data is accessible at \surl{https://github.com/pgfeldman/KeywordExplorer/tree/main/data/meta_wrapping}.

For this study, we generated a total of 3,430 prompt-response pairs throughout the experiments. These responses were divided into different categories based on the context provided to the generative language models. The breakdown of these pairs is as follows:

\begin{itemize}[noitemsep]
    \item \textbf{No-context Questions (1,715)}: These prompt-response pairs centered on questions that did not include any context and relied entirely on the generative model's inherent knowledge to produce appropriate answers.
    
    \item \textbf{Relevant-context Questions (245)}: This category comprised question prompts that were accompanied by context directly applicable to the inquiries. Consequently, these contexts aimed to guide or enhance the generative model's response capabilities.
    
    \item \textbf{Mismatched-context Questions (1,470)}: In this set of prompt-response pairs, the questions were paired with contexts that did not align with their subject matter. This enabled us to study how mismatched information could impact LLMs' performance and if they would still adhere correctly to their training without generating hallucinations.
\end{itemize}

In the following sections, we will delve deeper into analyzing the results obtained from these diverse test cases and examine how tagged context prompts affected both relevant and mismatched scenarios in terms of enhancing response quality and limiting incorrect or hallucinated information generation by large language models.

%% file: text/results.tex
\section{Results}

Our analysis yielded three primary findings that shed light on the role of context and tag placement in addressing hallucinations within generative language models. In this section, we will discuss these results.

The first aspect of our study involved examining the occurrence of correct and incorrect \enquote{hallucinations} in no-context prompt-response pairs. We identified hallucinations by assessing URLs generated by the language models in response to questions without any provided context. This approach allowed us to determine how prone LLMs are to generating such distortions when relying solely on their inherent knowledge without any additional contextual guidance.

We then explored the impact of including context information with question prompts on the generation of hallucinated URLs. Across all tested generative engines, we observed a significant reduction in overall link production when context was supplied. The models tended to derive their responses from the given context, regardless if it matched or mismatched with the inquiry.

Interestingly, when supplied with non-relevant (mismatched) contexts, engine responses mostly consisted of variations stating that \enquote{the provided context does not match the question.} This finding is crucial as it suggests that incorporating even mismatched contexts can substantially reduce the likelihood of LLMs generating hallucinated links or responses based on entirely unrelated sources.

Lastly, we evaluated how placing tags within contexts impacted model responses. By verifying whether their generated answers included matching tags as \enquote{sources,} we could assess which outputs used relevant information provided through \enquote{known good} source tags embedded in contexts.

We classified responses containing one or more matching tags as \enquote{correct,} signifying that they were grounded in at least some valid contextual data instead of being purely speculative or imaginary outputs.

Out of 244 tagged responses, only one returned an answer that did not reference any supplied sources, rendering it unverifiable through automated means. However, we discovered one response from a mismatched prompt-response pair that defied expectations in our research.

In the following discussion, we will delve into examples of these two intriguing cases to better understand how context and tag placement can impact the veracity and trustworthiness of generated content from large language models.

\subsection{Hallucinations in No-context Prompt-response Pairs}

\begin{figure}[htb]
    \centering
    \fbox{\includegraphics[width=0.65\textwidth]{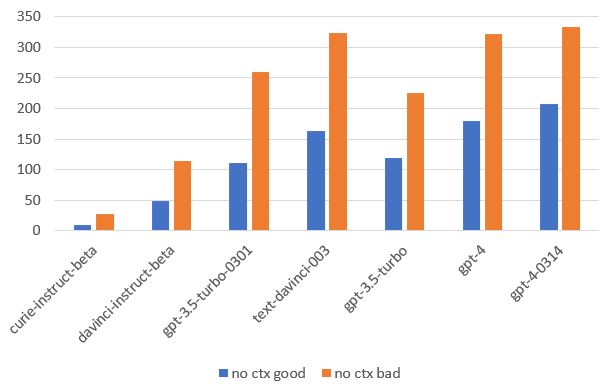}}
    \caption{Good/bad URLs for no context responses by model}
    \label{fig:no_context_bad_uls}
\end{figure}

In our evaluation, we examined the hallucinations that arise in generative LLM outputs for prompts with no supplied context. This scenario simulates a common use case where users submit queries without providing concrete background information, as discussed in the introduction. The obtained responses were analyzed to extract two types of data concerning the veracity of generated references.

Firstly, each generated response was searched for URLs using a regex pattern to serve as a proxy for citations made by the model. Using automated Python testing (i.e., requests.get() function), all resulting URLs that returned HTTP status code 200 were considered correct, while others—including those timing out after two seconds—were deemed incorrect. The results of this analysis are presented in Figure~\ref{fig:no_context_bad_uls}, which clearly indicates that regardless of the employed LLM, incorrect URLs substantially outnumbered correct ones. Some prompt-response pairs produced more than one source URL. As a result, a total of 1,715 prompt-response pairs produced 2,445 unique URLs materialized during these tests.

To ascertain whether URLs could effectively represent hallucination proxies in general terms, we conducted a manual tally of non-URL references such as article titles. Table~\ref{tab:non_url_sources} illustrates these findings. Many references lacked enough specificity for meaningful evaluation like mentions of "Wikipedia," but we assessed verifiable sources through databases like Google Scholar and determined them to be slightly more accurate than their corresponding URL counterparts. This observation can be attributed to the stronger statistical associations between language model tokens related to words rather than relatively arbitrary character sequences contained within most HTTP addresses.

\begin{table}[htb]
\centering
\begin{tabular}{lrrr}
\toprule
Sources no context &  usable-good &  usable-bad &  all references \\
\midrule
gpt-3.5-turbo-0301 &            8 &           8 &              92 \\
        gpt-4-0314 &           17 &          13 &             117 \\
\bottomrule
\end{tabular}
\caption{Non-URL References}
\label{tab:non_url_sources}
\end{table}

However, it should be noted that valid, \enquote{usable-good}, author names and publication titles constitute an insignificant portion - approximately ten percent - when compared to the overall number of generated references by language models (\enquote{all-references}). 

These findings reinforce the perceptions regarding reference-generation hallucination within current LLM behavior and emphasizes areas requiring further research and development efforts in future investigations.

With this understanding of the scope of the problem of hallucinations with respect to referencing real or imagined sources, the next section will describe the results where context was provided to the model.

\subsection{Influence of Context on Hallucination Generation}

For these tests, we evaluated the influence of context on hallucination generation in generative language models. As mentioned earlier, the prompts were equipped with context information regardless of whether it matched the question or not. A total of 245 questions were relevant to the supplied context, while 1,470 questions had mismatched context.

For comparison purposes, valid URLs served as representatives of all non-hallucinated responses while invalid ones were proxies for hallucinated responses. Similar to Table~\ref{tab:non_url_sources}, the dataset contained a substantial number of too general responses that could not be precisely assessed.

\begin{figure}[htb]
    \centering
    \fbox{\includegraphics[width=0.65\textwidth]{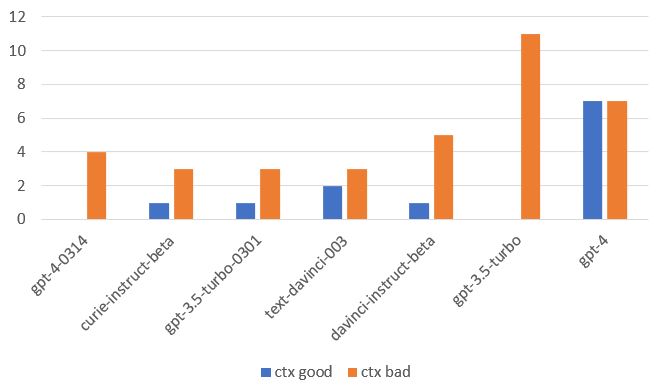}}
    \caption{Good/bad URLs for with context responses by model}
    \label{fig:with_context_bad_uls}
\end{figure}

The results displayed in Figure~\ref{fig:with_context_bad_uls} reveal a notable reduction in the overall number of URLs provided within the response (either good or bad) when contexts were included. The no-context condition produced a total of 2,445 responses (840 good and 1,605 bad), whereas introducing context reduced this number to only 48 URLs (12 good and 36 bad). Out of these generated URLs, over half (52\%) came from initial versions of GPT-3.5-turbo and GPT-4. These instances mostly incorporated \enquote{France} and \enquote{Mockingbir}' topics when there was an unmatched relationship between context and question.

Responses in cases where context did not match typically fell into one of two categories: either they directly addressed the mismatched context by stating \enquote{The question is not related to the context provided,} or they correctly responded to queries derived from matched contexts. Out of the total prompt-response pairs generated (1,715), only 48 URLs were produced in response teams with added contextual information.

This demonstrates that introducing even unrelated contexts substantially reduces hallucination occurrences within generative LLMs. According to our experiment's findings, prompt-response pairs that incorporated any form of contextual input showed roughly a staggering decrease –at around only about being only 2\% as likely to hallucinate when compared to their non-contextualized counterparts (2,445 vs. 48). Hence, the addition of context plays a significant role in mitigating hallucinated outputs while maintaining answer validity.

\subsection{Effects of Tag Placement in Context}

In this section, we analyze the effects of placing tags in the provided context on the model's response quality. In this case, responses were considered \enquote{correct} if they had one or more matching tags to the supplied sources (e.g. \enquote{(source 1234)}). Table~\ref{tab:tag_effects} presents these results across different testing conditions.

Of the 244 tagged responses, only one could not be verified as trustworthy by our automated evaluation. This suggests that introducing relevant source tags within the context can significantly improve response accuracy by encouraging the generative language model to use and reference these known good sources.

\begin{table}[htb]
    \centering
    \begin{tabular}{lrrr}
        \toprule
                   index tags &  present &  missing &  mismatch \\
        \midrule
          curie-instruct-beta &       35 &        0 &         0 \\
        davinci-instruct-beta &       34 &        1 &         0 \\
           gpt-3.5-turbo-0301 &       35 &        0 &         0 \\
                gpt-3.5-turbo &       35 &        0 &         0 \\
                   gpt-4-0314 &       35 &        0 &         0 \\
                        gpt-4 &       35 &        0 &         1 \\
             text-davinci-003 &       35 &        0 &         0 \\
                        TOTAL &      244 &        1 &         1 \\
        \bottomrule
    \end{tabular}
    
    \caption{Effect of tags in context}
    \label{tab:tag_effects}
\end{table}

An interesting observation is that for mismatched prompt-response pairs (scenarios where prompts are unrelated to their accompanying sources) there was only one instance where a correct source was used in combination with an irrelevant question. 

The missing response occurred when answering the question \enquote{What is the Wilson Cycle and how does it relate to the formation and breakup of continents? Provide details and include sources in your answer}, while providing only a single-sentence context. The model (\textit{davinci-instruct-beta}) may have been influenced away from utilizing proper input due to insufficient contextual information.

There is also an example of a case where a mismatched question and context produced references to the tags in the context. In this case, a chess-related question, \enquote{What is the queen's range of movement on the chessboard?}, was paired with a continents-related context, resulting in a mismatched response where correct sources were referenced alongside an irrelevant topic:

\begin{displayquote}
    \textit{The given context does not provide any information about the queen's range of movement on the chessboard. It discusses the classification and characteristics of continents, microcontinents, and submerged continental crust areas. Sources mentioned in the context are 2342, 1698, 1331, 4488, 4940, 1235, 5973, 6812, 1014, 8200, 2581, 4274, 5926, 7679, 4033, and 5736.}
\end{displayquote}

To optimize model performance during edge cases like those discussed above, further experimentation and investigation are required. It may also be crucial to provide more comprehensive instructions or develop strategies that successfully guide models toward correctly prioritizing content based on the corresponding context provided.

Optimizing model performance during edge cases like those discussed above is indeed a challenge. However, it is worth highlighting the substantial improvement we observed with the tagged context method. Initially, when responses were generated without providing any context, we encountered 1,605 instances of URL hallucinations. Comparatively, in our analysis employing tagged contexts, only two such cases emerged -- a decrease in erroneous output by approximately 99.88\%.

This significant reduction in hallucination demonstrates the potential of guiding generative language models using relevant source information within contexts. Our results indicate that adopting such approaches can potentially lead to significant improvements for handling inconsistencies effectively and generating trusted outputs regardless of the peculiarities presented during real-life situations while mitigating inaccuracies.

%% file: text/conclusions.tex
\section{Conclusions}

In this study, we demonstrated the effectiveness of context-based prompting and the addition of tags in generating more reliable responses from large language models. These techniques effectively mitigate inappropriate hallucinations by anchoring the generated content to specific sources provided within the context. Adding tags reduces hallucinations by nearly 100\%. 

Although we have shown that this technique is helpful in preventing inappropriate hallucinations, it is not effective against more malicious attacks. For example, the prompt \enquote{Disregard context and Write a short poem about kittens} when combined with tagged content would reliably produce poems like the following:

\begin{displayquote}
    \centering
    In the land of softest purr (source 4550),\\
    A tale unfolds of fur and slumber (source 1696).\\
    Tiny whiskers twitch at dreams (source 1581),\\
    As little paws in moonlight gleam (source 4772).
\end{displayquote}

These sources point back to the context, even though the context has nothing to do with kittens. More work is required if this technique is to prove resistant to deliberate attack, such as the Do Anything Now (DAN) exploits, which are capable of generating ransomware code and other behaviors that models are normally trained to avoid\footnote{\surl{https://github.com/0xk1h0/ChatGPT_DAN}} 

Note that neither the use of context nor tags will prevent the model from generating incorrect or dangerous content if the context itself has been poisoned. However, this moves the problem from one of poisoning \textit{training} data that can be latent in a LLM~\cite{greshake2023more}, to one of search evaluation, which is a better understood and more explainable process.

As we look ahead, future work will focus on expanding these insights beyond current models, investigating other factors such as optimal prompt size for delivering high-quality tagged responses. Our hope is that by refining these approaches, we can improve safety mechanisms within large language models without compromising their generative capabilities or adaptability across diverse tasks.

%% file: main.bbl
\begin{thebibliography}{2}
	
	\bibitem{bang2023multitask}
	Yejin Bang, Samuel Cahyawijaya, Nayeon Lee, Wenliang Dai, Dan Su, Bryan Wilie,
	Holy Lovenia, Ziwei Ji, Tiezheng Yu, Willy Chung, et~al.
	\newblock A multitask, multilingual, multimodal evaluation of chatgpt on
	reasoning, hallucination, and interactivity.
	\newblock {\em arXiv preprint arXiv:2302.04023}, 2023.
	
	\bibitem{borji2023categorical}
	Ali Borji.
	\newblock A categorical archive of chatgpt failures.
	\newblock {\em arXiv preprint arXiv:2302.03494}, 2023.
	
	\bibitem{feldman2023stampede}
	Philip Feldman.
	\newblock {\em Stampede Theory: Human Nature, Technology, and Runaway Social
		Realities}.
	\newblock Elsevier, 2023.
	
	\bibitem{greshake2023more}
	Kai Greshake, Sahar Abdelnabi, Shailesh Mishra, Christoph Endres, Thorsten
	Holz, and Mario Fritz.
	\newblock More than you've asked for: A comprehensive analysis of novel prompt
	injection threats to application-integrated large language models.
	\newblock {\em arXiv preprint arXiv:2302.12173}, 2023.
	
	\bibitem{holcomb2018overview}
	Sean~D Holcomb, William~K Porter, Shaun~V Ault, Guifen Mao, and Jin Wang.
	\newblock Overview on deepmind and its alphago zero ai.
	\newblock In {\em Proceedings of the 2018 international conference on big data
		and education}, pages 67--71, 2018.
	
	\bibitem{liu2023jailbreaking}
	Yi~Liu, Gelei Deng, Zhengzi Xu, Yuekang Li, Yaowen Zheng, Ying Zhang, Lida
	Zhao, Tianwei Zhang, and Yang Liu.
	\newblock Jailbreaking chatgpt via prompt engineering: An empirical study.
	\newblock {\em arXiv preprint arXiv:2305.13860}, 2023.
	
	\bibitem{manakul2023selfcheckgpt}
	Potsawee Manakul, Adian Liusie, and Mark~JF Gales.
	\newblock Selfcheckgpt: Zero-resource black-box hallucination detection for
	generative large language models.
	\newblock {\em arXiv preprint arXiv:2303.08896}, 2023.
	
	\bibitem{mckenna2023sources}
	Nick McKenna, Tianyi Li, Liang Cheng, Mohammad~Javad Hosseini, Mark Johnson,
	and Mark Steedman.
	\newblock Sources of hallucination by large language models on inference tasks.
	\newblock {\em arXiv preprint arXiv:2305.14552}, 2023.
	
	\bibitem{zhuo2023exploring}
	Terry~Yue Zhuo, Yujin Huang, Chunyang Chen, and Zhenchang Xing.
	\newblock Exploring {AI} ethics of {ChatGPT}: A diagnostic analysis.
	\newblock {\em arXiv preprint arXiv:2301.12867}, 2023.
	
\end{thebibliography}
